\documentclass[letterpaper]{article} 
\usepackage{aaai2026}  
\usepackage{times}  
\usepackage{helvet}  
\usepackage{courier}  
\usepackage[hyphens]{url}  
\usepackage{graphicx} 
\usepackage{caption} 
\usepackage{natbib}  

\usepackage{algorithm}
\usepackage{algorithmic}
\usepackage{bm}

\usepackage{amsmath}
\usepackage{amsfonts}

\usepackage{multirow}
\usepackage{booktabs}
\usepackage{makecell}

\title{NADIR: Differential Attention Flow for Non-Autoregressive Transliteration in Indic Languages}

\author{
    Lakshya Tomar\textsuperscript{\rm 1},
    Vinayak Abrol\textsuperscript{\rm 2},
    Puneet Agarwal\textsuperscript{\rm 1}
}
\affiliations{
    \textsuperscript{\rm 1}RocketFrog AI\\
    \textsuperscript{\rm 2}CSE Department, Indraprastha Institute of Information Technology, Delhi\\
    lakshya.tomar@rocketfrog.ai, abrol@iiitd.ac.in, puneet@rocketfrog.ai
}

\begin{document}
\maketitle

\begin{abstract}
In this work, we argue that not all sequence-to-sequence tasks require the strong inductive biases of autoregressive (AR) models. Tasks like multilingual transliteration, code refactoring, grammatical correction or text normalization often rely on local dependencies where the full modeling capacity of AR models can be overkill, creating a trade-off between their high accuracy and high inference latency. While non-autoregressive (NAR) models offer speed, they typically suffer from hallucinations and poor length control. To explore this trade-off, we focus on the multilingual transliteration task in Indic languages and introduce \textbf{NADIR}, a novel NAR architecture designed to strike a balance between speed and accuracy. NADIR integrates a Differential Transformer and a Mixture-of-Experts mechanism, enabling it to robustly model complex character mappings without sequential dependencies. NADIR achieves over a 13$\times$ speed-up compared to the state-of-the-art AR baseline. It maintains a competitive mean Character Error Rate of \textbf{15.78\%}, compared to \textbf{14.44\%} for the AR model and \textbf{21.88\%} for a standard NAR equivalent. \textbf{Importantly}, NADIR reduces Repetition errors by \textbf{49.53\%}, Substitution errors by \textbf{24.45\%}, Omission errors by \textbf{32.92\%} and Insertion errors by \textbf{16.87\%}. This work provides a practical blueprint for building fast and reliable NAR systems, effectively bridging the gap between AR accuracy and the demands of real-time, large-scale deployment.
\end{abstract}

\section{Introduction}
    Transformers~\cite{vaswani2023attentionneed} suffer from attention noise~\cite{ye2025differentialtransformer}, assigning irrelevant attention scores to unimportant context, which can degrade model performance. This issue is further amplified in Non-Autoregressive (NAR)~\cite{gu2018} settings due to reduced contextual information, making it difficult for NAR models to match the performance of their Autoregressive (AR)~\cite{sutskever2014} counterparts. This gap is largely due to insertions, substitutions, omissions, and repetitions errors, which we collectively refer to as \textbf{NAR Hallucinations}. However, we argue that not all tasks benefit from the strong sequential bias of AR models. Tasks such as multilingual transliteration, code refactoring, and grammatical correction often rely more on local dependencies. In this work, we show that reducing attention noise and incorporating a Mixture-of-Experts (MoE)~\cite{foundationalMOE} architecture enables NAR models to achieve competitive performance while being significantly more efficient. To validate our approach, we focus on Indic transliteration as a representative task.

Indic scripts include a diverse family of writing systems such as Devanagari (used in Hindi, Marathi, Sanskrit), Bengali, and Punjabi/Gurmukhi, which have been deeply rooted in South Asian cultures for centuries. Today, more than \textbf{1.6 billion people} use these scripts in various languages. A key linguistic challenge with these scripts is transliteration, the process of converting text from one script to another while preserving pronunciation. Transliteration differs from translation: It maps the sounds of a word, not its meaning. 
Transliteration is inherently challenging due to: (a) ambiguity in character mappings, including many-to-one, one-to-many, and many-to-many relationships, (b) phonetic variability, where different words in a non-English language may be transliterated into the same Roman word, and (c) homophones and phonological constraints, where similar sounds are represented by different characters depending on context. Thus, the context of surrounding characters is essential for accurate transliteration, which is why AR models typically deliver superior performance.

While there is a growing shift from AR to NAR models in related tasks, techniques such as Knowledge Distillation~\cite{gong2022knowledgetransferdistillationautoregressive}, Iterative Refinement~\cite{lee2020iterativerefinementcontinuousspace}, and CTC Loss~\cite{nar_ctc}, as detailed in Related Work, have been employed to mitigate the accuracy loss incurred from abandoning AR approaches. However, to our knowledge, this issue remains unresolved for transliteration tasks. In this work, to address this need, we propose \textbf{\textit{NADIR} (Non-Autoregressive Differential Intelligent Router)}, a multilingual deep learning architecture inspired by the differential transformer and Mixture-Of-Expert, capable of transliterating \textbf{180.1k} words in just \textbf{2 minutes and 55 sec.} ($\approx 1005$ words/second). In contrast, the current state-of-the-art IndicXLIT model~\cite{d:aksharantar} requires \textbf{38 minutes and 50 sec.} ($\approx 77$ words/ second). Also, Our model achieves a competitive mean CER of 15.78\% (STD: 5.67\%) versus IndicXLIT's 14.44\% (STD: 4.35\%). The key contributions of our work are as follows:

\begin{figure*}[t]
  \centering
    \includegraphics[scale=.41]{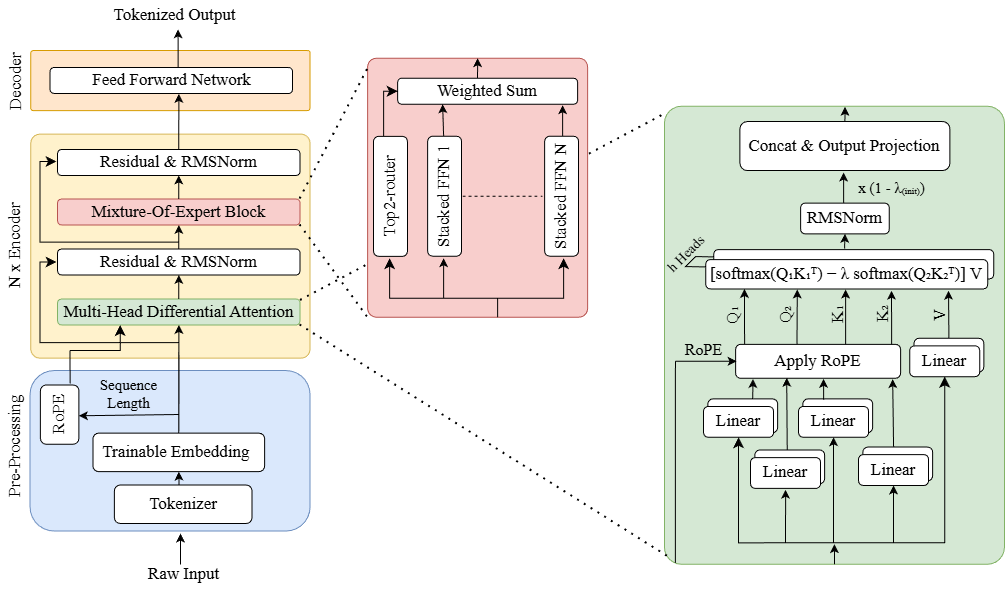}
  \caption{\textit{NADIR:} Architecture Overview showing pre-processing, multiple encoder and non-autoregressive decoder blocks. On the right, we show an expanded view of the encoder block that is built upon differential transformer layers combined with MoE-based routing, enabling efficient and expressive context modelling.}
  \label{fig:nadir-architecture}
\end{figure*}

\begin{enumerate}
    \item We present an analysis of failure modes that arise when switching from autoregressive to non-autoregressive models, collectively referred to as \textit{NAR Hallucinations}. 
    \item \textbf{First Non-Autoregressive Transliteration}: State-of-the-art transliteration models have very slow decoding speed, making them impractical for large-scale or latency-sensitive applications. We address this gap by introducing \textit{NADIR} a novel transliteration model designed for extreme throughput without sacrificing accuracy.
    \begin{enumerate}
        \item \textbf{Scalable Transliteration via \textit{NADIR}}: To the best of our knowledge, this is the first work to address the hallucination problem in non-autoregressive NLP models for transliteration. 
        \item Our approach integrates Differential Attention Mechanism~\cite{ye2025differentialtransformer}, which significantly reduces repetition, substitution and omission errors by minimizing attention noise. Complementing this, the MoE module further reduces insertion and remaining repetition and substitution errors by enabling dynamic, token-specific computation — as confirmed by our ablation study.
    \end{enumerate}
    
\end{enumerate}

\textbf{In summary, the research question we addressed:}
Can reducing attention noise and incorporating MoE help NAR models capture context effectively without auto-regression?

Next, we position our work in the context of prior research in the Related Work section. We then present a detailed description of the proposed \textit{NADIR} architecture, along with the underlying intuition behind its design. This is followed by a brief overview of the datasets used and a comprehensive explanation of the training setup, including model parameters and optimization strategies. In the Experimental Results section, we report empirical comparisons against strong AR and NAR baselines, highlighting \textit{NADIR}’s performance across multiple error categories. We also include an ablation analysis to assess the individual contributions of key architectural components. Finally, we conclude with a summary of our findings and outline directions for future work.

\section{Related Work}

NAR models have emerged as a promising alternative to traditional AR methods for sequence-to-sequence tasks, offering significant inference speed-ups. However, this parallelism often comes at the cost of `hallucinations' errors such as token repetitions, substitutions, omissions and insertions, due to the model's inherent conditional independence assumption. Previous research has explored several strategies to mitigate these issues, each with its own set of trade-offs.

An early and prominent line of work involves leveraging an AR `teacher' model to guide the training of an NAR `student' through knowledge distillation\cite{gong2022knowledgetransferdistillationautoregressive}. While such methods have demonstrated improved performance, this dependency on a pre-trained AR model introduces significant complexity, increases overall training time, and moves away from a truly end-to-end parallel training paradigm. Other approaches have focused on modifying the loss function or model architecture. The Connectionist Temporal Classification (CTC) loss, particularly explored in tasks like speech recognition and Chinese-Braille translation~\cite{nar_ctc}, facilitates alignment but is limited by its assumption of monotonic alignment and conditional independence between output tokens. These assumptions are often violated in complex transliteration and translation scenarios. Similarly, early architectures like FlowSeq~\cite{ma:2019} attempted to address NAR limitations by incorporating a small auxiliary network for explicit length prediction. However, an inaccurate length prediction can trigger cascading errors, destabilizing the entire generation process.

To address the core issue of conditional independence, NAR paradigms have evolved, primarily in the field of machine translation with the most successful approaches being iterative refinement. Instead of generating the entire sequence in a single pass, these models iteratively edit a draft translation over a fixed number of steps. Seminal models like Mask-Predict~\cite{ghazvi2019} use a conditional masked language model objective to re-predict subsets of tokens that it is least confident about. Similarly, the Levenshtein Transformer~\cite{Jiatao2019} explicitly trains the model to perform editing operations like insertion and deletion. While these methods significantly close the quality gap with AR models, they trade some of the latency gains for higher accuracy. Another advanced technique involves incorporating latent variables to guide the parallel decoding process. Models like the Disentangled Context Transformer~\cite{kasai2020} and Glancing Transformer~\cite{qian2021} learn latent alignments or fertilities that provide a soft structural blueprint for the output, improving coherence without full sequential conditioning. A more recent work has also explored energy-based models~\cite{lifu2020} to improve the quality of the generated sequences.

Despite these advancements in general sequence-to-sequence modeling, the specific application of advanced NAR techniques to multilingual transliteration remains underexplored. Transliteration presents a unique challenge: it requires strict phonetic and orthographic accuracy but is more constrained and localized than open-domain translation. The overhead of iterative refinement may be unnecessary, while the simple conditional independence of basic NAR models is insufficient. This highlights a clear need for an architecture that can achieve a robust balance—capturing local dependencies with high fidelity while retaining maximum parallelism. Our work, \textit{NADIR}, is designed to fill this gap by integrating a differential attention mechanism and expert Gating Network~\cite{foundationalMOE}, tailored specifically for the demands of high-throughput, multilingual transliteration. In contrast to existing works, the proposed NAR model enables fully parallel and efficient training/decoding. We demonstrate that token-level cross-entropy loss is enough to train the model \& this avoids CTC’s rigid alignment and independence assumptions, leading to better fluency and reordering capability. In addition, NADIR bypasses explicit length prediction, removing a major source of instability in prior NAR systems.

\section{NADIR (Non-Autoregressive Differential Intelligent Router)}

We introduce \textbf{NADIR}, a NAR multilingual transliteration model designed to address hallucination problems in transliteration, while maintaining the speed and competitive performance compared to SOTA AR models. As illustrated in Figure~\ref{fig:nadir-architecture}, the overall pipeline begins with a preprocessing stage that applies a tokenizer along with learnable token and rotary positional embeddings (RoPE)~\cite{rope}. Our core innovation lies in the stacked encoder blocks, each built upon Differential Transformer~\cite{ye2025differentialtransformer} layers combined with MoE-based routing, enabling efficient and expressive context modeling. A lightweight MLP-based non-autoregressive decoder generates output characters (to the target script vocabulary) in parallel using the refined encoder representations.

\subsubsection{Multi-head Differential Attention:}
In NAR, the absence of the sequential inductive bias makes it difficult for the standard attention mechanism to focus on the most relevant input tokens. This often leads to noisy attention maps, because of which we observed issues like token insertions, substitutions, omissions, or repetitions. To mitigate this, one can use \textbf{Differential Attention Mechanism}, that computes the difference between normalized-softmax attention scores and regulates it using a learnable parameter $\lambda$. Formally, given a $d$-dim input sequence \( X \in \mathbb{R}^{M \times d} \) \& length $M$, the query and key projections are each partitioned into two components ($Q_1, Q_2$ and $K_1, K_2$) as
\[
    [Q_1; Q_2] = XW^Q; \quad [K_1; K_2] = XW^K; \quad  V = XW^V
\]
The attention output is computed by modulating the primary attention score with the secondary one, controlled by a learnable scalar $\lambda$:
\begin{equation}
    \label{eq:diffattn}
    \text{DiffAttn}(X) = \left( \text{S}\left( \frac{Q_1 K_1^\top}{\sqrt{d}} \right) - \lambda \, \text{S}\left( \frac{Q_2 K_2^\top}{\sqrt{d}} \right) \right)V
\end{equation}
Here, $S(\cdot)$ is the softmax function, and $W^Q, W^K, W^V$ are the projection matrices. To ensure stable training, the dynamic modulator $\lambda$ is parameterized using learnable vectors ($\bm{\lambda}_{q}, \bm{\lambda}_{k}$) and an initial bias $\lambda_{\text{init}}$:
\[
    \lambda = \exp(\bm{\lambda}_{q_1} \cdot \bm{\lambda}_{k_1}) - \exp(\bm{\lambda}_{q_2} \cdot \bm{\lambda}_{k_2}) + \lambda_{\text{init}}
\]
A key yet important detail in NADIR is the use of \textbf{RMSNorm}~\cite{rms}, which empirically performs consistently better than the \textbf{GroupNorm}~\cite{groupNorm} in the conventional differential attention block.

\subsubsection{Mixture-of-Expert Module:}
Error analysis of the differential transformer based model (without MoE) demonstrated better performance on languages that had more training data than others. This observation led us to hypothesize that a single shared Feed-Forward Network (FFN) might not effectively capture the diversity of scripts for all languages. As an initial step, we replaced a single FFN in each layer with a set of small FFNs, referred to as ``expert'' and manually routed tokens based on linguistic or regional knowledge. While this hardcoded routing showed promising results, particularly for low-resource languages, its effectiveness was likely limited by the small capacity of each expert. One workaround could be scaling expert size based on data availability per cluster, but such heuristics break generality and are not scalable. To address this, we transitioned to a Mixture-of-Experts (MoE) framework, where routing is learned dynamically. A learned router FFN assigns tokens to experts, which potentially enables flexible and context-aware specialization. This MoE design showcases a bit of robustness in multilingual settings - as shown in the ablation study. Formally MoE layer contains $M$ expert feed-forward networks, denoted as $\{E_1, E_2, ..., E_M\}$. For each input token representation \( x \in \mathbb{R}^{d} \), a trainable gating network $G(x)$ computes routing probabilities over all experts:
\[
p_i = \frac{\exp(g_i(x))}{\sum_{j=1}^{M} \exp(g_j(x))}, \quad \text{for } i = 1, \dots, M
\]
where $g_i(x)$ is the logit score for the $i^{\text{th}}$ expert. In our setup, we adopt \textbf{Top-2 routing}~\cite{moe}, where only the two experts with the highest gating scores are selected for each input token. Let $i$ and $j$ be the indices of the top-2 experts, then the final output is computed as:
\[
\text{MoE}(x) = p_i \cdot E_i(x) + p_j \cdot E_j(x)
\]

\subsection{Implicit Sequence Termination and Training Objective}
A fundamental challenge for NAR models is determining when to terminate generation. Unlike AR models, which naturally predict an end-of-sequence (\texttt{[EOS]}) token, NAR models generate all tokens in parallel. While prior work often resorts to a separate, complex length-predictor network, this can introduce instability.

We adopt a more direct approach by enabling the NAR model to perform implicit length prediction. To achieve this, we append an \texttt{[EOS]} token to every target sequence during training. The key innovation lies in our loss computation: it is calculated only over the token positions up to and including the first predicted \texttt{[EOS]} token, thereby explicitly encouraging the model to learn when to stop generation. During inference, the model generates tokens up to a fixed maximum length, and we extract the output up to the first position where the model predicts the \texttt{[EOS]} token. This strategy elegantly compels the model to learn sequence boundaries without an auxiliary network, though its effectiveness is highest in tasks with clearly defined termination points, e.g., it will not generalize to tasks with highly ambiguous or variable-length outputs. To support this mechanism, we employ a composite loss function, which is a weighted sum of two key components, namely; 

\begin{enumerate}
    \item \textbf{Token-Level Cross-Entropy Loss} to ensure local prediction accuracy. Let \( \hat{\mathbf{Y}} \in \mathbb{R}^{B \times T \times V} \) be the predicted logits for a batch of size \( B \), with sequence length \( T \) and vocabulary size \( V \), and \( \mathbf{Y} \in \mathbb{N}^{B \times T} \) the ground-truth targets, the loss is computed as
    \[
    \mathcal{L}_{\text{token}} = \frac{1}{B \cdot T} \sum_{b=1}^{B} \sum_{t=1}^{T_b} -\log P(Y_{b,t} \mid \hat{Y}_{b})
    \]

    \item \textbf{Load-Balancing Loss~\cite{foundationalMOE} } to ensure uniform expert utilization in our Mixture-of-Experts (MoE) layers. Let \( \mathbf{G} \in \mathbb{R}^{B \times M} \) denote the gating probabilities over \( M \) experts for each input in the batch, the loss is computed as:
    \[
    \mathcal{L}_{\text{load}} = M \cdot \sum_{e=1}^{M} \left( \frac{1}{B} \sum_{b=1}^{B} G_{b,e} \right)^2
    \]
\end{enumerate}
The final training objective is a weighted sum of the above:
\[
\mathcal{L}_{\text{total}} = \alpha \mathcal{L}_{\text{token}} + \beta \mathcal{L}_{\text{load}}
\]
where \( \alpha, \text{and } \beta \) are scalar hyperparameters controlling the contribution of each loss component.

\subsection{Why Differential Transformer Helps in Transliteration}
In a standard multi-head attention~\cite{vaswani2023attentionneed} layer, the outputs of all heads are aggregated, typically through concatenation followed by a linear projection. However, this aggregation can lead to ambiguous representations. If some heads capture features for the correct output while others capture a simpler, incorrect alternative, the final vector is muddled as a blend of both.  The Differential Transformer enhances this process. Instead of a simple sum, it computes a weighted sum over the attention heads, operating on the differences between adjacent head states. This ``differential softmax'' reintroduces local context and selectivity. This approach offers two key advantages for NAR transliteration. First, it encourages smoother, more stable layer-wise refinement of token representations. The model learns how much of the ``difference'' to apply at each step, preventing abrupt changes that cause hallucinations. Secondly, it provides fine-grained, content-aware control over how features evolve through the layers. This is crucial for transliteration, where local phonetic consistency is key, as it aligns better with the task's monotonic structure. For instance, consider the input `ksha' (devnagari word). Additive heads might capture the general `complex consonant cluster' feature, while a subtractive head can learn to specifically identify the most common simplification, like the feature for `ka'. The final representation becomes: \textit{[(cluster feature) - $\lambda$ (`ka' feature)]}, actively carving the `ka' ambiguity out of the representation, leaving a sharper, more precise vector that points decisively to `ksha'.

\section{Experimental Setup}
This section we provide experimental protocol and the dataset used in the experimental study.

\begin{table*}[th]
    \centering
    \resizebox{\textwidth}{!}{
    \begin{tabular}{l|cc|cc|cc||cc|cc|cc}
    \hline
    \multirow{3}{*}{\textbf{Language}} & \multicolumn{6}{c||}{\textbf{Roman $\rightarrow$ Indic}} & \multicolumn{6}{c}{\textbf{Indic $\rightarrow$ Roman}} \\ \cline{2-13}
    & \multicolumn{2}{c|}{\textbf{CER} $\downarrow$} & \multicolumn{2}{c|}{\textbf{WAcc} $\uparrow$} & \multicolumn{2}{c||}{\textbf{InfT (sec.)} $\downarrow$} & \multicolumn{2}{c|}{\textbf{CER} $\downarrow$} & \multicolumn{2}{c|}{\textbf{WAcc} $\uparrow$} & \multicolumn{2}{c}{\textbf{InfT (sec.)} $\downarrow$} \\ \cline{2-13}
    & IndicXLIT & NADIR & IndicXLIT & NADIR & IndicXLIT & NADIR & IndicXLIT & NADIR & IndicXLIT & NADIR & IndicXLIT & NADIR \\ \hline
    Telugu (Tel)      & 11.03 & \textbf{9.15} & 58.66 & \textbf{66.92} & 136.22 & \textbf{10.42} & \textbf{11.97} & 12.50 & \textbf{41.85} & 41.43 & 152.82 & \textbf{10.37} \\
    Maithili (Mai)    & \textbf{11.67} & 12.60 & \textbf{61.24} & 58.56 & 165.48 & \textbf{5.34} & 15.79 & \textbf{15.09} & 37.24 & \textbf{39.48} & 64.65 & \textbf{5.62} \\
    Bengali (Ben)     & \textbf{15.01} & 18.06 & \textbf{52.29} & 46.29 & 189.94 & \textbf{14.11} & \textbf{20.87} & 21.34 & \textbf{25.09} & 24.16 & 212.87 & \textbf{14.05} \\
    Nepali (Nep)      & \textbf{9.93} & 10.80 & 65.18 & \textbf{64.76} & 36.00 & \textbf{4.03} & \textbf{8.31} & 8.90 & \textbf{65.76} & 64.74 & 39.78 & \textbf{4.32} \\
    Assamese (Asm)    & \textbf{12.29} & 16.65 & \textbf{52.8} & 43.11 & 59.72 & \textbf{5.23} & \textbf{14.8} & 15.67 & \textbf{40.27} & 37.92 & 65.81 & \textbf{5.80} \\
    Malayalam (Mal)   & 14.29 & \textbf{12.85} & 43.31 & \textbf{55.39} & 165.48 & \textbf{12.77} & \textbf{13.76} & 14.08 & 33.28 & \textbf{33.3} & 186.32 & \textbf{12.73} \\
    Urdu (Ur)         & \textbf{18.38} & 21.13 & \textbf{43.58} & 37.08 & 202.84 & \textbf{14.94} & \textbf{26.35} & 27.07 & \textbf{17.58} & 16.08 & 226.42 & \textbf{14.62} \\
    Hindi (Hin)       & \textbf{11.45} & 13.47 & \textbf{58.18} & 54.01 & 131.26 & \textbf{10.14} & \textbf{15.17} & 15.80 & \textbf{39.73} & 38.56 & 146.69 & \textbf{10.29} \\
    Sindhi (Sid/Sin)  & \textbf{19.44} & 22.15 & \textbf{47.86} & 38.87 & 76.90 & \textbf{6.71} & \textbf{23.32} & 24.12 & \textbf{23.54} & 22.32 & 84.36 & \textbf{7.34} \\
    Tamil (Tam)       & 17.7 & \textbf{12.40} & 28.72 & \textbf{60.95} & 149.69 & \textbf{11.82} & \textbf{17.77} & 18.14 & \textbf{30.97} & 30.52 & 166.39 & \textbf{11.89} \\
    Manipuri (Man)    & \textbf{9.59} & 14.05 & \textbf{65.43} & 53.28 & 47.57 & \textbf{4.46} & \textbf{13.51} & 18.06 & \textbf{49.5} & 39.39 & 52.54 & \textbf{5.13} \\
    Sanskrit (San)    & 15.21 & \textbf{14.88} & 49.77 & \textbf{51.56} & 56.56 & \textbf{5.24} & \textbf{11.71} & 12.50 & \textbf{44.44} & 43.59 & 62.56 & \textbf{5.54} \\
    Gujarati (Guj)    & \textbf{11.02} & 12.31 & \textbf{58.94} & 54.88 & 251.54 & \textbf{17.97} & \textbf{21.58} & 21.71 & \textbf{22.01} & 21.38 & 285.41 & \textbf{17.70} \\
    Punjabi (Pan)     & \textbf{18.47} & 20.78 & \textbf{43.39} & 38.42 & 141.14 & \textbf{11.29} & \textbf{20.78} & 21.76 & \textbf{30.11} & 28.07 & 158.38 & \textbf{11.19} \\
    Konkani (Kok)     & \textbf{14.46} & 15.65 & \textbf{51.63} & 48.67 & 51.07 & \textbf{4.98} & \textbf{17.89} & 19.41 & \textbf{31.48} & 28.34 & 56.38 & \textbf{5.33} \\
    Kashmiri (Kas)    & \textbf{27.46} & 34.32 & \textbf{25.51} & 15.35 & 86.30 & \textbf{6.88} & \textbf{26.71} & 29.60 & \textbf{19.3} & 16.00 & 96.76 & \textbf{7.14} \\
    Boro (Brx)        & \textbf{14.13} & 16.20 & \textbf{52.17} & 47.90 & 35.61 & \textbf{4} & \textbf{13.25} & 14.91 & \textbf{48.96} & 44.69 & 39.10 & \textbf{4.31} \\
    Odia (Ori)        & \textbf{17.55} & 18.05 & \textbf{41.56} & 39.75 & 37.61 & \textbf{4.31} & \textbf{12.33} & 13.50 & \textbf{47.66} & 44.58 & 41.67 & \textbf{4.49} \\
    Kannada (Kan)     & 9.37 & \textbf{8.04} & 62.02 & \textbf{68.36} & 149.71 & \textbf{11.53} & \textbf{9.68} & 10.38 & \textbf{44.23} & 43.03 & 167.00 & \textbf{11.19} \\
    Marathi (Mar)     & \textbf{10.29} & 12.09 & \textbf{62.46} & 58.43 & 158.91 & \textbf{12.21} & \textbf{16.42} & 16.67 & \textbf{32.99} & 32.43 & 176.73 & \textbf{12.4} \\ \hline
    Mean    & 14.44 & 15.78 & 51.23 & 50.13 & 116.48 & 8.95 & 16.59 & 17.56 & 36.29 & 34.5 & 124.18 & 9.07 \\
    Std. Dev.  & 4.35 & 5.67 & 10.87 & 12.21 & 62.71 & 4.14 & 5.10 & 5.3 & 11.75 & 11.5 & 71.03 & 3.95 \\
    \hline
    \end{tabular}
    }
\caption{Comprehensive comparison of baseline (IndicXLIT) and proposed model (\textit{NADIR}) for transliteration in both Roman $\leftrightarrow$ Indic directions. Metrics include Character Error Rate (CER), Word-level Accuracy (WAcc), and Inference Time (IntT). For each metric, the better result per language is in bold}
\label{tab:main_res}
\end{table*}

\subsection{Dataset}
We evaluate the performance of various transliteration models on the Aksharantar dataset \cite{d:aksharantar}, which is the largest open-source parallel dataset available for transliteration in multiple Indian languages. This dataset includes parallel word-level mappings between English and 21 Indic languages and vice versa. 
Aksharantar dataset includes parallel corpora across 21 Indian languages. The largest training sets are for Malayalam (4.1M), Tamil (3.2M), Kannada (2.9M), and Telugu (2.4M). Languages like Gujarati, Hindi, and Bengali each contribute over 1M training samples. Several mid- to low-resource languages, including Urdu (699k), Konkani (612k), Panjabi (514k), and Oriya (346k), are also represented. Low-resource languages such as Kashmiri (46k), Sindhi (59k), Bodo (35k), Manipuri (106k) included to test generalization. The total dataset comprises \textbf{24.8 million} training, \textbf{129.6k} validation, and \textbf{180.1k} test samples.

\subsection{Model Training Setup}
\label{sec:exp}
We trained two models, one for each direction, i.e., Roman-to-Indic and Indic-to-Roman. Both models were trained for 100 epochs using 4 encoder layers. Each encoder layer contains 8 attention heads and 5 experts, Each expert comprises a stack of 
$\text{FFN}(\text{embed\_dim} \times \frac{\text{expert\_dim}}{2}) \rightarrow \text{GELU} \rightarrow 
\text{FFN}(\frac{\text{expert\_dim}}{2} \times \text{expert\_dim}) \rightarrow \text{GELU} \rightarrow 
\text{FFN}(\text{expert\_dim} \times \text{embed\_dim})$. Where expert\_dim and embed\_dim are 512, 768 respectively. We use the AdamW~\cite{optimizer} optimizer with a learning rate of $1 \times 10^{-3}$ and a weight decay of $1 \times 10^{-3}$, along with a linear learning rate scheduler~\cite{huggingface} where 15\% of the total steps are allocated for warmup, with Dropout~\cite{srivastava2014} of 0.1 and Capacity Factor~\cite{foundationalMOE} of 1.25. The best performance on \textit{NADIR} was observed when hyperparameters \( \alpha \), and \( \beta \) were set to \textbf{0.8}, and \textbf{0.2}, in our experiments.

The total number of parameters in \textit{NADIR} is approximately \textbf{27 million}. Training is conducted on two NVIDIA RTX 3090 GPUs, inference is performed using a single GPU with a batch size of 8192.

\section{Experimental Results}

To evaluate our hypotheses, we present a comparative analysis of performance and inference speed for the proposed \textit{NADIR} model and IndicXLIT, the current state-of-the-art autoregressive model for transliteration, under various experimental scenarios.  We systematically vary key components of the model to understand their individual and combined effects on transliteration performance.

\subsection{Main Result}

In Table~\ref{tab:main_res}, we present a detailed comparison between the proposed model \textit{NADIR} and the baseline model IndicXLIT in both transliteration directions (Roman$\leftrightarrow$Indic) across 20 Indian languages. We evaluate the models using three metrics: \textbf{Character Error Rate (CER~$\downarrow$)}, \textbf{Word Accuracy (WAcc~$\uparrow$)}, and \textbf{Inference Time (InfT~$\downarrow$)}.

\begin{table}[t!]
    \centering
    \begin{tabular}{c|c|c}
    \hline
    \textbf{Ground Truth} & \textbf{NAR Output} & \textbf{AR Output} \\
    \hline
        direktorrao & direkt\textbf{ara}rao &  direktorrao  \\
        mushtaidi & mus\textbf{htaadad} & mushtaidi  \\
        undannadi & und\textbf{hanna} & undannadi \\
        samvardhita & sam\textbf{babababadhadha}ta & samvardhita  \\
    \hline
    \end{tabular}
    \caption{Examples showing hallucinations in NAR outputs compared to AR outputs. All outputs are in Roman script for readability.}
    \label{tab:ar-vs-nar}
\end{table}

\begin{table*}[th!]
\centering
\begin{tabular}{l|c|c|c|c}
\toprule
\textbf{Model} & \textbf{Insertion} & \textbf{Substitution} & \textbf{Omissions} & \textbf{Repetition} \\
\hline
Standard NAR              & 28,454 & 72,127 & 37,769 & 6,313 \\
\hline
Diff NAR                  & 27,682 & 57,349 & \textbf{23,453} & 4,126 \\
Gain over Standard NAR&2.71\% & 20.49\% & \textbf{37.90\%}        &  34.64\%     \\
\hline
Diff MoE NAR              & \textbf{23,654} & \textbf{54,494} & 25,334 & \textbf{3,186} \\
Gain over Standard NAR  &  \textbf{16.87\%}      &  \textbf{24.45\%}      & 32.92\%       &  \textbf{49.53\%}     \\
Gain over Diff NAR  & 14.55\%       &    4.98\%    & -8.02\%       & 22.78\%      \\
\hline
\end{tabular}
\caption{Hallucination Error Breakdown count across NAR Variants (lower is better). Gains are shown with respect to the baseline (Standard NAR) and the previous row.}
\label{tab:hallucination_breakdown}
\end{table*}

\begin{table*}[th!]
    \centering
    \begin{tabular}{c|c|c}
    \hline
    \textbf{Ground Truth} & 
    \makecell{\textbf{Encoders using}\\\textbf{Standard Attention}} & 
    \makecell{\textbf{Encoders using}\\\textbf{Differential Attention}} \\
    \hline
    direktorrao & direkt\textbf{ara}rao & direkt\textbf{or}rao \\    
    mushtaidi & mushta\textbf{adad} & mushta\textbf{idi} \\
    undannadi & und\textbf{hanna} & und\textbf{annadi} \\
    mononayonpotro & mono\textbf{yoyonpot} & mono\textbf{nayonpotro} \\
    mononoyonprotyaashi & mono\textbf{yoyprapr}yaashi & mono\textbf{noyonprot}yaashi \\
    \hline
    \end{tabular}
    \caption{Examples showing hallucinations in NAR outputs and how Differential Attention reduces such errors.}
    \label{tab:standard-vs-diff}
\end{table*}

\subsubsection{Indic$\rightarrow$Roman direction:} On average, across all 20 languages in the Indic$\rightarrow$Roman transliteration task, \textit{NADIR} achieves a CER of \textbf{17.56}, which is comparable to IndicXLIT's \textbf{16.59}. The average WAcc of \textit{NADIR} is \textbf{34.5\%}, closely matching IndicXLIT's \textbf{36.29\%}. More importantly, \textit{NADIR} offers a substantial efficiency advantage—reducing the mean InfT from \textbf{124.18 seconds} to just \textbf{9.07 seconds}, representing an order-of-magnitude speedup.

\subsubsection{Roman$\rightarrow$Indic direction} Across all 20 languages, \textit{NADIR} achieves an average CER of \textbf{15.78}, closely aligned with IndicXLIT's \textbf{14.44}. The WAcc of \textit{NADIR} is \textbf{50.13\%}, slightly below IndicXLIT's \textbf{51.23\%}. Once again, \textit{NADIR} demonstrates a significant reduction in inference latency, bringing down the mean InfT from \textbf{116.48 seconds} to just \textbf{8.95 seconds}. Notably, despite similar overall averages, \textit{NADIR} outperforms IndicXLIT in both CER and WAcc for \textbf{5 out of the 20 languages}, underscoring its robustness across diverse scripts.

We also consider the difference between metrics for \textit{IndicXLIT} and \textit{NADIR} models, denoted by $\Delta$. For the Indic$\rightarrow$Roman direction, the mean and standard deviation of $\Delta$ CER were \textbf{-0.96} and \textbf{-0.2}, respectively, indicating slightly better character-level accuracy for IndicXLIT. The corresponding values for $\Delta$ WAcc were \textbf{1.80} and \textbf{0.25}. Inference Time, however, shows a substantial gain for \textit{NADIR}, with a mean $\Delta$ InfT of \textbf{115.11 seconds} and a standard deviation of \textbf{67.08}. The Roman$\rightarrow$Indic direction follows a similar trend: $\Delta$ CER had a mean and standard deviation of \textbf{-1.34} and \textbf{-1.32}, while $\Delta$ WAcc values were \textbf{1.1} and \textbf{-1.34}. Inference time gains remained consistent, with a mean $\Delta$ InfT of \textbf{107.53 seconds} and a standard deviation of \textbf{58.57}.

These results demonstrate that while \textit{NADIR} slightly trails IndicXLIT in accuracy, it delivers comparable overall performance with a dramatic reduction in inference time, making it a highly practical choice for real-time and large-scale deployment scenarios.

\begin{table}[ht]
    \centering
    \begin{tabular}{l|c|c}
        \toprule
        \textbf{Model} & \textbf{mean CER} & \textbf{mean WAcc} \\
        \midrule
        Standard Attention & 21.88 & 38.98 \\
        Differential Attention & 16.12 & 46.89 \\
        Diff-Attn + MoE & \textbf{15.78} & \textbf{50.13} \\
        \bottomrule
    \end{tabular}
    \caption{Average CER and Accuracy across all languages for different model variants.}    
    \label{tab:avg-performace}
\end{table}

\subsection{Ablation Study}
In our experiments with the standard encoder-based NAR model, we observed all four types of NAR Hallucination, namely, character insertions, substitutions, omissions, and repetitions. Table~\ref{tab:ar-vs-nar} presents representative failure cases, highlighting these Hallucinations, by comparing AR and NAR predictions.

We measured the extent of NAR hallucinations across the outputs of three model variants: a)~Standard NAR, b)~Differential Transformer-based NAR (Diff NAR), and c)~Differential Transformer with Mixture-of-Experts (Diff MoE NAR). Insertion, substitution, and omission errors were quantified using the \texttt{editdistance} algorithm. Repetition errors were further categorized into three types: Insert Repeat (repeated spans not present in the ground truth), Substitute Repeat (repeated spans that replace expected content but are not part of the ground truth), and Valid Repeat (spans present in the ground truth but repeated more times than necessary). We considered character spans ranging from bigrams to four-grams and counted each distinct repeated span only once, regardless of its frequency. The results of these measurements are summarized in Table~\ref{tab:hallucination_breakdown}. 
        
\subsubsection{Impact of Differential Attention Mechanism}
Switching from the standard to the Differential Attention Mechanism helps alleviate the \textbf{NAR Hallucination} problem discussed in the previous section. We hypothesize that the subtraction operation in Equation~\ref{eq:diffattn} effectively suppresses attention noise, thereby minimizing the influence of irrelevant context that often leads to hallucinations, particularly in the absence of strong autoregressive signals. This hypothesis is supported by the quantitative results presented in Table~\ref{tab:hallucination_breakdown}, where the Differential NAR model demonstrates significant improvements in addressing substitutions, omissions, and repetitions, though it shows limited impact on reducing insertions. Qualitative examples illustrating these improvements are provided in Table~\ref{tab:standard-vs-diff}, and they align closely with the statistical trends observed.

\begin{table}[ht]
\centering
\begin{tabular}{c|c|c}
\toprule
\textbf{Ground Truth} &
\makecell{\textbf{Encoders}\\\textbf{without MOE}} &
\makecell{\textbf{Encoders}\\\textbf{with MOE}} \\
\midrule
ṭoṛe & ṭōṛ & ṭoṛ\textbf{e} \\
mahama & maham\textbf{am} & maham\textbf{a} \\
anukarana & anukara\textbf{nn}a & anukara\textbf{n}a \\
emmelayelapai & \textbf{mla}lapai & \textbf{emmelaye}lapai \\
dasakare & dasaka\textbf{rere} & dasaka\textbf{re} \\
\bottomrule
\end{tabular}

\caption{Examples showing how MoE enhances script-specific accuracy over Differential Attention-only models. All text is in Roman script for clarity.}
\label{tab:Diff-vs-MOE-DIFF}
\end{table}

\subsubsection{Impact of Mixture-of-Experts (MoE) with Differential Attention Mechanism}
While Differential Attention significantly mitigates \textbf{NAR hallucinations}, models trained solely with this mechanism continue to struggle with the introduction of unwanted characters, particularly insertions (as shown in Table~\ref{tab:hallucination_breakdown}). Moreover, there remains substantial room for improvement in other hallucination types.

To address these limitations, we incorporate a Mixture-of-Experts (MoE) module within each encoder layer. This addition yields a notable reduction in insertion errors and also shows meaningful improvements in handling repetitions. While the MoE-NAR variant achieves a modest gain in reducing substitution errors, it does introduce a slight increase in omission errors. Table~\ref{tab:Diff-vs-MOE-DIFF} provides qualitative examples where the MoE-enhanced architecture successfully corrects errors that the Differential Attention Mechanism alone fails to resolve. Additionally, as shown in Table~\ref{tab:hallucination_breakdown}, the incorporation of the MoE module leads to a reduction in Insertion, Substitution, and Repetition errors by \textbf{14.55\%}, \textbf{4.98\%}, and \textbf{22.78\%}, respectively albeit with an approximate \textbf{8\%} increase in omission errors.

\subsubsection{Effect of Batch Size on Inference Time} 
\begin{figure}[t]
    \centering
    \includegraphics[width=0.97\linewidth]{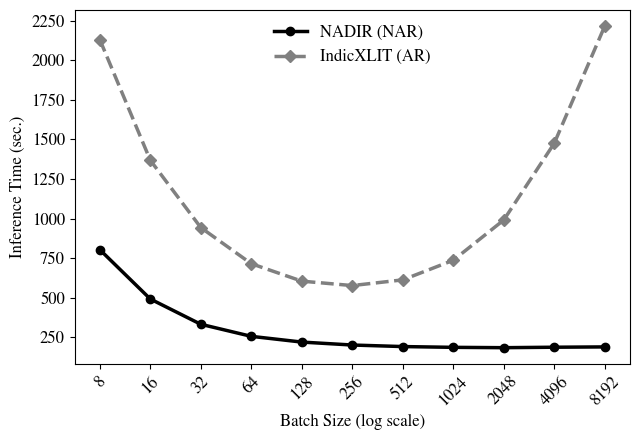}
    \caption{
        Inference Time vs. Batch Size (log scale). \textit{NADIR} achieves significantly lower inference latency compared to the AR-based IndicXLIT, especially at large batch sizes.}
    \label{fig:inference-time}
\end{figure}

Figure~\ref{fig:inference-time} illustrates the inference time (in seconds) of \textit{NADIR} and IndicXLIT models across varying batch sizes on a log scale. As expected, both models benefit from increased batch sizes up to a point, after which inference time starts to rise due to memory and hardware constraints. \textit{NADIR} consistently outperforms IndicXLIT across all batch sizes, demonstrating significantly lower latency and better scalability. Notably, while IndicXLIT exhibits a sharp U-shaped curve with a narrow optimal batch window, \textit{NADIR} achieves near-optimal performance over a much wider range, highlighting its robustness and suitability for high-throughput deployment.

\subsection{Summary}        
To summarize the findings, transitioning from AR to NAR introduces consistent errors, collectively referred to as NAR Hallucination. The Differential Transformer effectively addresses most of these issues and serves as the primary contributor to performance improvements. Additional enhancements, such as incorporating Mixture-of-Experts (MoE), further help mitigate difficult edge cases where even the Differential Transformer falls short. Table~\ref{tab:avg-performace} reports the mean CER and WAcc for each model variant. 

\section{Conclusion}
In this work, we revisit the design space of non-autoregressive  models for sequence-to-sequence tasks where local dependencies dominate, using multilingual transliteration as a representative case. We present \textbf{NADIR}, a novel NAR architecture that combines a Differential Transformer with a Mixture-of-Experts framework to tackle key challenges such as hallucinations, poor length control, and loss of linguistic fidelity.

Our findings demonstrate that NADIR delivers significant improvements over standard NAR baselines while achieving an order-of-magnitude reduction in inference time compared to autoregressive (AR) models. Crucially, NADIR reduces repetition, substitution, omission, and insertion errors substantially, thereby closing the accuracy gap with AR models without inheriting their computational overhead.

This work provides both a conceptual and empirical foundation for designing high-throughput, accurate NAR systems that are well-suited for real-time applications in resource-constrained or large-scale multilingual settings. Looking ahead, we believe that the principles underlying NADIR differential attention, expert specialization, and hallucination aware evaluation can be extended to a broader class of structured generation tasks beyond transliteration. We also aim to extend NADIR to tasks beyond transliteration, with a focus on incorporating recent advancements in Mixture-of-Experts into the architecture.

\section*{Acknowledgements}
This work was supported by Infosys Foundation via the Infosys Centre for Artificial Intelligence, Indraprastha Institute of Information Technology Delhi, and in part by the Nebius Research Grant.

\bibliography{aaai2026}

\end{document}